\pdfoutput=1

\documentclass[11pt]{article}

\usepackage[]{EACL2023}

\usepackage{times}
\usepackage{latexsym}

\usepackage[T1]{fontenc}

\usepackage[utf8]{inputenc}

\usepackage{microtype}

\usepackage{inconsolata}

\usepackage{epsfig}
\usepackage{amsmath}
\usepackage{amssymb}
\usepackage{booktabs}
\usepackage[]{graphicx}
\usepackage[export]{adjustbox}
\usepackage{tikz}
\usepackage{floatrow}
\usepackage{arydshln}
\usepackage{xcolor,colortbl}
\usepackage{adjustbox}
\usepackage{floatrow}
\usepackage{tabularx}
\usepackage{todonotes}
\usepackage{multirow}
\usepackage{soul}
\usepackage{wrapfig,lipsum,booktabs}
\usepackage{subcaption}
\usepackage{comment}
\usepackage{color}

\usepackage{float}
\floatstyle{plaintop}
\usepackage{url}
\usepackage{blindtext}
\usepackage{enumitem}
\usepackage{amssymb}
\usepackage{multirow}
\usepackage{arydshln}
\usepackage{fnpct}
\newcommand{\dataset}{\texttt{HVVMemes}}
\newcommand{\model}{\texttt{VECTOR}}
\newcommand\sbullet[1][.5]{\mathbin{\vcenter{\hbox{\scalebox{#1}{$\bullet$}}}}}
\definecolor{deepgreen}{rgb}{0.01, 0.75, 0.24}
\usepackage{inconsolata}

\newcommand\blfootnote[1]{%
  \begingroup
  \renewcommand\thefootnote{}\footnote{#1}%
  \addtocounter{footnote}{-1}%
  \endgroup
}
\title{Characterizing the Entities in Harmful Memes:\\Who is the Hero, the Villain, the Victim?}

\author{Shivam Sharma$^{1,4}$$^{*}$, Atharva Kulkarni$^{2}$$^{*}$, Tharun Suresh$^{3}$, Himanshi Mathur$^{3}$, \\\textbf{Preslav Nakov$^{5}$, Md. Shad Akhtar$^3$ and Tanmoy Chakraborty$^1$}\\
  $^1$Indian Institute of Technology Delhi, India  $\sbullet[.75]$
  $^2$Carnegie Mellon University, USA\\
  $^3$Indraprastha Institute of Information Technology Delhi, India $\sbullet[.75]$ $^4$Wipro AI Labs, India\\
  $^5$Mohamed bin Zayed University of Artifcial Intelligence, UAE \\
  \small\texttt{\{shivam.sharma, tanchak\}@ee.iitd.ac.in}, \small\texttt{atharvak@cs.cmu.edu}, \small\texttt{preslav.nakov@mbzuai.ac.ae}, \\ \small\texttt{\{tharun20119, himanshi18037, shad.akhtar\}@iiitd.ac.in}
  }

\begin{document}
\maketitle
\begin{abstract}
Memes can sway people's opinions over social media as they combine visual and textual information in an easy-to-consume manner. As they can turn viral, it it crucial to infer their intent and potentially associated harm to take timely measures. A common problem associated with meme comprehension lies in detecting the entities referenced and characterizing the role of each of these entities. Here, we aim to understand whether the meme glorifies, vilifies, or victimizes each entity it refers to. To this end, we address the task of {\em role identification of entities in harmful memes}, i.e., detecting who is the `hero', the `villain', and the `victim' in the meme, if any. We use \dataset, a memes dataset on US Politics and COVID-19, released recently as part of the CONSTRAINT@ACL-2022 shared-task. It contains memes, entities referenced, and their associated roles: hero, villain, victim, and other. We further design \model\ (Visual-semantic role dEteCToR), a robust multi-modal framework for the task, which integrates entity-based contextual information in the multi-modal representation and we compare it to several standard unimodal (text-only or image-only) or multi-modal (image+text) models. The experimental results show that our proposed model achieves an improvement of 4\% over the best baseline and 1\% over the best competing stand-alone submission from the shared task. 
Finally, we highlight the challenges encountered in addressing the complex task of semantic role labeling of memes. 
\end{abstract}

\section{Introduction}

\begin{figure}
    \centering
    \includegraphics[width=0.8\columnwidth]{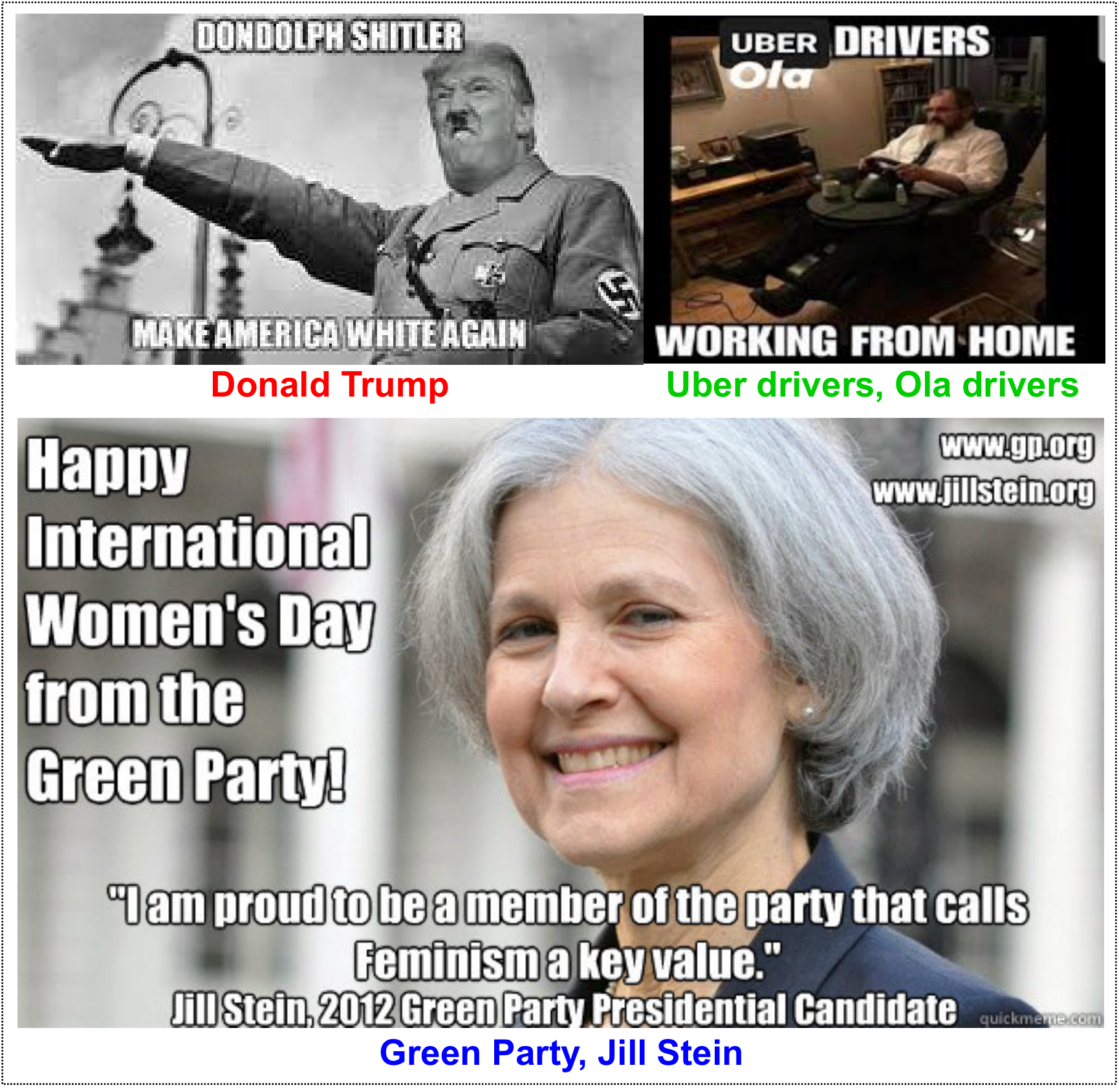}
    \caption{Examples of \texttt{\textcolor{blue}{heroes}}, \texttt{\textcolor{red}{villains}} and \texttt{\textcolor{deepgreen}{victims}}, as portrayed within memes.}
\label{fig:meme_hvv_eg}
\end{figure}

\blfootnote{$^*$Equal contribution}
Social media have emerged as a conducive medium for information exchange. Yet, their democratic nature has fostered unabated dissemination of hate speech \cite{MacAvaney2019Hate}, misinformation \cite{wu2019misinformation}, fake news \cite{ALDWAIRI2018215}, propaganda \cite{da-san-martino-etal-2020-semeval}, and other harmful content.

Such information manifests itself in various ways, and more recently, in the form of \emph{memes}. Though typically intended to burlesque and lampoon world events, political outlook, or daily routine, an ostensibly innocuous meme can readily become a multi-modal cause of distress with a dexterous blend of images and texts. Due to their viral nature and ability to circumvent censorship \cite{mina2014batman}, social media instigators and hatemongers are increasingly using memes as a powerful medium for disseminating spiteful content. Therefore, investigating the dark side of memes has risen as a pertinent research problem both in industry and in academia \cite{sharma-etal-2020-semeval,pramanick-etal-2021-detecting}.

{\bf Motivation.} While there have been many studies that have analyzed memes through the lens of emotions \cite{sharma-etal-2020-semeval}, sarcasm \cite{kumar2019sarc}, hate speech \cite{zhou2021multimodal,kiela2020hateful}, misinformation \cite{zidani2021memes}, offensiveness \cite{suryawanshi-etal-2020-multimodal}, and harmfulness \cite{pramanick-etal-2021-detecting,pramanick-etal-2021-momenta-multimodal}, there has been much less focus on analyzing semantic roles present within memes. 

Understanding these roles is critical for comprehending memes and narrative framing of various social entities. Caricaturing these entities with nefarious motives can lead to misinformation propagation, social calumny, and hatred towards minorities. In addition to the dark portrayals, memes sometimes depict the sorrowful state of certain entities, illustrate their heroism, etc. Consider the memes in Fig.~\ref{fig:meme_hvv_eg}. The meme in Fig.~\ref{fig:meme_hvv_eg} (a) portrays Jill Stein and the Green Party as \emph{heroes} for their feminist views. Fig.~\ref{fig:meme_hvv_eg} (b) draws a comparison between Adolf Hitler and Donald Trump, thus portraying the latter as a \emph{villain} for his anti-immigration views. Fig.~\ref{fig:meme_hvv_eg} (c), on the other hand, depicts the plight of Ola and Uber drivers, who are out of work and \emph{victims} of the lockdowns due to the COVID-19 pandemic. Thus, through depictions of heroism, villainy, and victimization, memes act as an alluring means to spread entity-relevant information and opinions.

{\bf Challenges.} Despite growing interest in analyzing memes, identifying the underlying connotations for the entities framed therein remains a challenging task \cite{sharma2022findings}. Memes are obscure due to their highly cryptic semantics, and satirical content \cite{sabat2019hate}. Moreover, categorizing the entities as a hero, a villain, or a victim requires real-world, contextual, temporal, spatial, and commonsense knowledge, which makes the task highly complex and subjective even for humans. Therefore, off-the-shelf multi-modal models that stand out well on conventional visual-linguistic tasks often flounder for memes as they are presumably inept to comprehend and capture the veiled information and multi-modal nuances present in a meme \cite{kiela2020hateful}. 

{\bf Our contributions.}  We propose a powerful approach to tackle the novel task of identifying the roles (the hero, the villain, and the victim) of the entities present in a meme. We model the problem as a role identification task and we report the results for several unimodal and multi-modal baselines to benchmark the task (and the dataset) and to assess its feasibility. We then proffer \model\, a vision and commonsense enriched version of DeBERTa \cite{he2020deberta} for the task at hand. As meme text often contains satirical content, which tends to contradict the meme image, it is necessary to consider mutual information from both modalities. Also, meme content is often stated in a non-obvious way, necessitating commonsense and world knowledge.

Thus, our \model\ attempts to infuse the relevant visual and commonsense knowledge with linguistic representations. Using ConceptNet \cite{speer2017conceptnet}, we generate an entity-relevant knowledge graph to represent commonsense knowledge relevant to the meme. Moreover, we use a distinct multi-modal information fusion strategy based on Optimal Transport. We appropriate Optimal Transport-based Kernel Embedding (OTKE) \cite{mialon2020trainable} for cross-modal correspondence. This technique by \citet{mialon2020trainable} marries the concepts of optimal transport theory with kernel techniques to provide robust and adaptable cross-modal adaptation. Our qualitative analysis underscores the importance of vision and commonsense knowledge integration, as \model\ outperforms several competitive baselines.

Our contributions can be summarized as follows:\footnote{The source code is available at \footnotesize{\url{http://github.com/LCS2-IIITD/VECTOR-Visual-semantic-role-dEteCToR}}.}\footnote{The dataset can be downloaded from the official shared-task page: \footnotesize{\url{http://codalab.lisn.upsaclay.fr/competitions/906}}.}

\begin{enumerate}[noitemsep,nolistsep]
    \item \textbf{Benchmarking \dataset:} We benchmark the \dataset\ via ten baselines with various unimodal and multi-modal systems.
    \item \textbf{Multi-modal system for identifying the hero, the villain, and the victim:} We develop \model\ (Visual-semantic role dEteCToR), a knowledge enriched multi-modal system that integrates entity-based knowledge in the multi-modal representations.
    \item \textbf{Extensive evaluation:} We report sizeable gains as part of our examination of \model\ compared to state-of-the-art models and the shared task submissions. 
    \item \textbf{Detailed Analysis:} Along with the ablation investigations, we provide detailed qualitative and quantitative analysis.
\end{enumerate}

\section{Related Work}


\paragraph{\bf Online Harmfulness.}
Due to the exponential rise of harmful content on various social media platforms, the research community has piqued its curiosity toward related studies. Some of them are based on online trolling \cite{ortiz2020_troll,cook2018troll},
 cyber-bullying \cite{choudhury21cyberbully,Kowalski2014bullying},
cyber-stalking \cite{ulbeh2011cyberstalk}, and hate speech \cite{MacAvaney2019Hate,zhang2018hate}.
Other studies characterize the correlation of racial and ethnic discrimination in the online and in the offline worlds \cite{Relia_Li_Cook_Chunara_2019}. \citet{chengtroll2017} examined the psycho-sociological outlook of online users toward online trolling behavior analysis. Few noteworthy investigations include characterizing homophily for self-harm due to eating disorders \cite{Chancellor2016ED,Wang2017ED} 
using logistic regression and snow-ball sampling, and suicide-ideation \cite{Burnap2015suicide,cao-etal-2019-latent} via linguistic, structural, affective and socio-psychological features. For a significant period, most of these studies have been dominated by text-oriented investigation while obscuring knowledge about other modalities.    

\paragraph{\bf Characterising Online Targets.}
Another research direction focuses on aspects such as relevance, stance, hate speech, sarcasm, and dialogue acts within hateful exchanges on Twitter in conventional and multi-task settings \cite{zain2017,Gautam_Mathur_Gosangi_Mahata_Sawhney_Shah_2020,ousidhoum-etal-2019-multilingual}. \citet{zain2018neural} addressed it by proposing neural networks with word embeddings. In contrast, the aspect-based sentiment was studied while addressing data sparsity, classification accuracy, and sarcastic content identification \cite{Zainuddin2019HateCO}. \citet{shvets-etal-2021-targets} demonstrated the efficacy of a generic concept extraction module for detecting the targets of hate speech. A few other studies on characterizing targets in harmful communication  \cite{sap-etal-2020-social,mathew2020hatexplain}
addressed social bias and hate speech explainability for targeted protected categories. \citet{ma-etal-2018-joint} used a hierarchical stack bidirectional gated recurrent units to detect targets and associated sentiments. A similar objective was studied in \cite{mitchell-etal-2013-open} but was formulated as sequence tagging in low-resource settings.  \citet{silva2016analyzing} used sentence structure to capture hate speech targets on social media to address detection and prevention. Most of these studies either focused on one primary designated target or emphasized detecting the association of sentiment while ignoring the affective spectrum. As observed in the literature \cite{shvets-etal-2021-targets}, such approaches may not generalize well across domains. 

\paragraph{\bf Studies on Memes.}
A significant influx of memes from online fringe communities, such as Gab, Reddit, and 4chan, to mainstream platforms, such as Twitter and Instagram, resulted in a massive epidemic of intended harm \cite{Zannettou2018}. 
Conventional visual features alongwith multimodal associativity was explored towards detecting memes in \cite{CURwebist20,memenonmeme}. 
Several datasets capturing offensiveness \cite{suryawanshi-etal-2020-multimodal}, hatefulness \cite{kiela2020hateful,gomez2019exploring}, and harmfulness \cite{pramanick-etal-2021-momenta-multimodal}, have been curated. Detecting memetic harmfulness and targeted categories are discussed in \cite{pramanick-etal-2021-momenta-multimodal}.
Commonsense knowledge \cite{9582340}, web entities, racial cues \cite{pramanick-etal-2021-momenta-multimodal,karkkainen2019fairface}, and other external cues have also been explored for detecting offense, harm, and hate speech in memes. Participatory events like the Facebook Hateful Meme Challenge \cite{kiela2020hateful} have laid a strong foundation for community-level initiatives for detecting hate speech in memes. As part of this challenge, several interesting approaches utilising meta information, attentive interactions, and adaptive loss are attempted in the multimodal setting \cite{das2020detecting,sandulescu2020detecting,zhou2020multimodal,lippe2020multimodal}.
Most of these efforts either address the detection tasks at various levels for harmfulness; see a recent survey \cite{Survey:2022:Harmful:Memes} or design \textit{ensemble} techniques lacking cost-optimality. However, as per our knowledge, no stand-alone approach reliably addresses the fine-grained task of understanding the roles of specific entities referred to within memes. 
We intend to address these aspects by seeking a robust and generalizable multimodal framework.

\begin{table}[t!]
    \centering
    \label{tab:dataset_summary}
    \resizebox{0.9\columnwidth}{!}{%
        \begin{tabular}{ccccccc}
            \toprule
            \multirow{2}{*}{\textbf{Domain}} & \multirow{2}{*}{\textbf{\# Memes}} & \multicolumn{5}{c}{\textbf{\# Entity References}} \\ \cmidrule(lr){3-7}
            &  & Hero & Villain & Victim & Other & Total \\ \midrule
            COVID-19 & 3381 & 200 & 747  & 407 & 3065 & 4419\\ 
            US Politics & 3552 & 288 & 1641 & 544 & 3242 & 5715\\ \bottomrule
        \end{tabular}}
    \caption{Summary of statistics for \# memes and entities referenced within them in \dataset\ \cite{sharma2022findings}. Original train/val/test split ratio of 80:10:10 ($\%$) for memes was preserved.}
    \label{tab:dataset_summary}
\end{table}

\section{Dataset}
\label{sec:dataset}
We employ \dataset, a dataset released as part of CONSTRAINT@ACL-2022. It contains English memes on two topics: $3,552$ memes about COVID-19 (C) and $3,381$ memes related to US Politics (P). The dataset primarily captures connotative roles: hero, villain, and victim, for different entities referenced within memes. 
\begin{figure*}[tbh]
    \centering
    \includegraphics[width=0.85\textwidth]{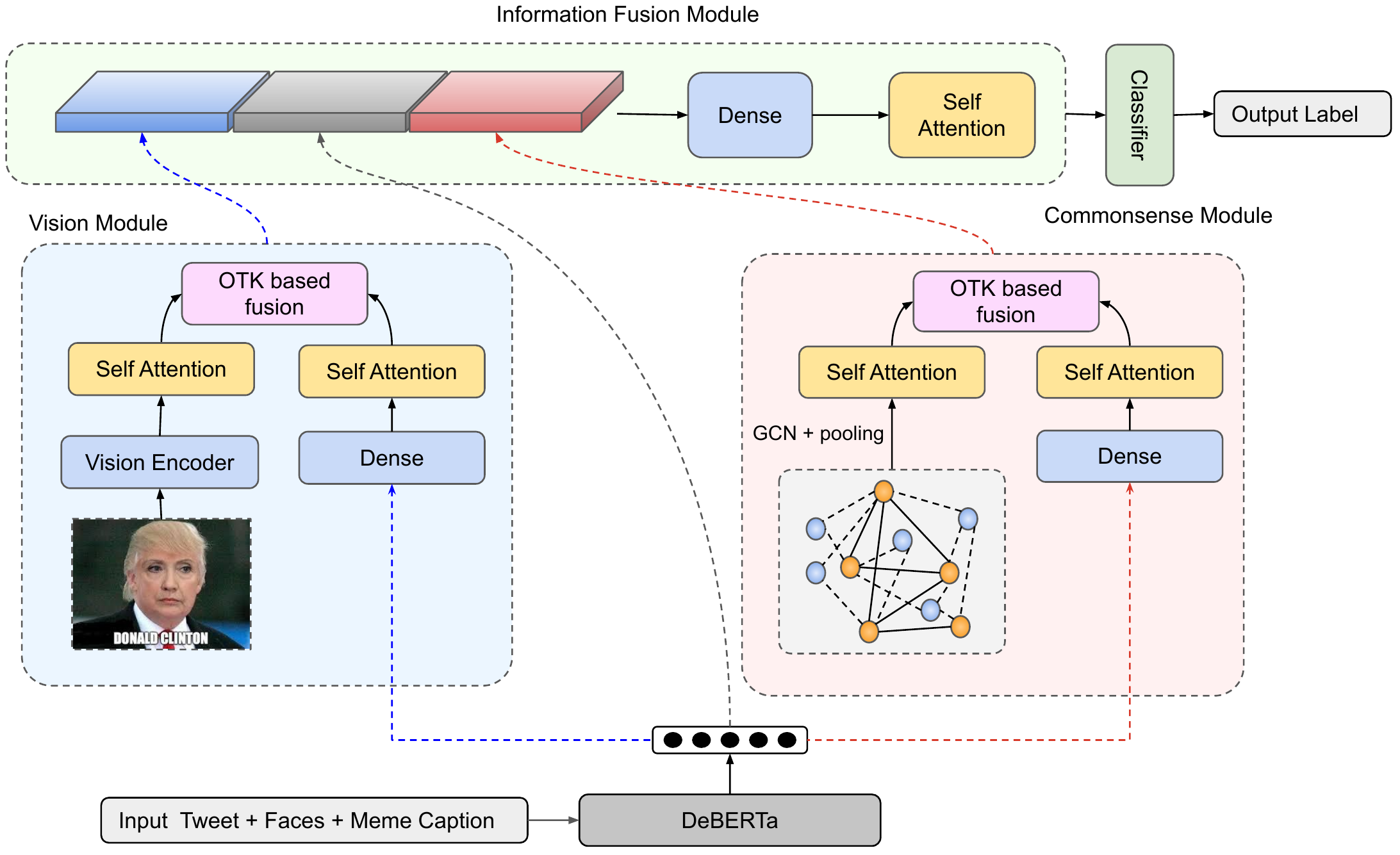}
    \caption{Schematic diagram of \model. The \textit{vision module} (left) consists of a ViT based image encoder and \textit{commonsense module} (right) leverages GCN-based commonsense knowledge graph embedding. \textit{Information fusion} module (top) fuses corresponding outputs toward final classification. OTK:  Optimal Transport-based Kernel.}
\label{fig:model_arch}
\end{figure*}
Table \ref{tab:dataset_summary} shows a summary of \dataset \footnote{For additional details, please refer to the shared-task paper \cite{sharma2022findings}.}. In general, most of the entities referenced in the memes do not have any connotation associated (C: $3,065$, P: $3,242$) and are categorised as \textit{other}. Amongst the key categories under consideration, \textit{villain} has the most candidate references (C: $747$, P: $1,641$), followed by \textit{victim} (C: $407$, P: $544$), and finally \textit{hero} (C: $200$, P: $288$), within a total of $3,381$ and $3,552$ memes for COVID-19 and US Politics, respectively. This highlights the realistic trends on social media.

\section{Proposed Approach}

This section outlines our proposed model \model\ (Visual-semantic role dEteCToR) and its varied components. As previously noted, role detection for memetic entities is challenging and requires real-world, contextual, and commonsense knowledge. Thus, we propose a neuro-symbolic approach that integrates commonsense-enriched modelling via graph (KG) structure into the language modelling-based architecture \cite{ZHANG202114}. KG's can be considered as discrete symbolic knowledge, which we leverage along with multimodal neural modelling. As shown in Fig.~\ref{fig:model_arch}, \model\ houses two primary sub-modules. The Vision Module leverages cross-modal interaction between the visual-linguistic signals to grasp optimal contextual information. The Commonsense Module integrates commonsense cues through an entity-based knowledge graph. 
Lastly, the Information Fusion Module coalesces the information obtained via attention-based fusion. In the following subsections, we go over the specifics of each module.

\paragraph{Text Module:}
We use DeBERTa \cite{he2020deberta} as our backbone model as it gives the best results amongst the text-only baselines (see Table~\ref{tab:baseline}). Aside from the text encoded in the meme, additional verbal information can be gleaned from memes. Evidence by \citet{blaier-etal-2021-caption} suggests that utilizing meme captions improves hateful meme identification results. Furthermore, additional cues such as the person, the location, and the entities present in the meme are helpful for downstream tasks. Thus, we use such ancillary information along with the OCR text. For image captioning, we make use of the recently released OFA model \cite{wang2022ofa}. For face identification, we use the same technique as the one by \citet{kun-etal-2022-logically}. The OCR text, the entity name, the image caption, and the identified face \textit{labels} are concatenated and passed to the DeBERTa model. We take the final layer representation $Z \in \mathbb{R}^{l \times d}$ to fuse information from other sub-modules.

\paragraph{Vision Module:}
Meme contents often contain contradicting text and image pairs. Therefore, it is required to incorporate information from these modalities to understand memes. Instead of using the traditional cross-modal attention to facilitate interaction between the two modalities, we utilize an optimal transport-based kernel interaction \cite{mialon2020trainable}. To begin with, we use a Vision Transformer \cite{dosovitskiy2020image} for generating the image representations $E_m \in \mathbb{R}^{l_m \times d_m}$. The text and the image representations undergo dimensionality reduction by non-linear transformation, followed by a self-attention layer \cite{NIPS2017_3f5ee243} as given by equations \ref{eq:att-feature-image}. This spawns vectors $Z'_m, E'_m \in \mathbb{R}^{l \times d'}$. We concatenate these two vectors and pass them to the Optimal Transport-based Kernel Embedding (OTKE) layer to bring about cross-modal interaction. It transforms the feature vectors to a Reproducing Kernel Hilbert Space (RKHS) \cite{bookAlain} followed by a weighted pooling scheme using weights determined by the transport plan between the set and a trainable reference. Such a fusion technique provides a theoretically grounded adaptive vector for the task. This yields vector $Z_m \in \mathbb{R}^{l \times d'}$, given by equation \ref{eq:OTKE-image} below:

\vspace{-0.7em}
\begin{small}
\begin{align}
  Z'_m &=& S\left(\frac{Z Z^T}{\sqrt{d}}\right)Z \quad E'_m &=& S\left(\frac{E_m E_m^T}{\sqrt{d_m}}\right)E_m  \label{eq:att-feature-image} \\
  Z_m &=& OTKE([Z'_m: E'_m])\label{eq:OTKE-image}
\end{align}
\end{small}

\paragraph{Commonsense Module:}
Due to their cryptic nature, identifying the intent of a meme is challenging. Their satirical and non-obvious way of conveying a message often requires commonsense comprehension. Thus, in our commonsense module, we generate a commonsense knowledge graph based on the entities in the meme. Similarly to \cite{9582340}, we extract all the nouns and noun phrases in the meme OCR and the meme caption. We extract commonsense relation pairs having confidence $>2$ from ConceptNet \cite{speer2017conceptnet} for all the noun chunks from memes. Noun chunks, commonsense entities, and the meme entity in question form the nodes of the commonsense graph. Each noun chunk from the OCR text is connected. The same applies to the noun chunks in the meme caption. A special aggregator token connects OCR and caption-based nodes. Thus, for each entity $e_{ij} \in E_i = \{e_{i1}, e_{i2}, \ldots , e_{in}\}$, we have a commonsense graph $G_{ij}=(V_{ij},E_{ij})$, where $V_{ij}$ and $E_{ij}$ are the nodes and the edges of the graph. 

The graph having edges between various entities, noun chunks, and nouns, being undirected, represents a generic ``connectivity''. Therefore, an edge between nodes A and B indicates that they have some association. By doing this, we wanted to capture the commonsense-based 'proximity' of different entities/concepts within the vectorized space for common sense concepts. $V_{ij}$ represents a set node (or vertices) constituting a commonsense graph corresponding to each entity.

For each node in the commonsense graph, we generate an embedding of size $d_g$ using the last layer representations from DeBERTa. In order to facilitate the interaction between the nodes, the commonsense graph goes through two rounds of graph convolutions \cite{kipf2016semi} followed by a max-pooling operation to spawn an aggregated graph embedding $E_g \in \mathbb{R}^{d'}$. Similarly to the vision module, the textual representations $Z$ and the commonsense graph representation $E_g$ undergo non-linear transformation and dimensionality reduction followed by self-attention \cite{NIPS2017_3f5ee243}. This generates contextual vectors $Z'_g, E'_g \in \mathbb{R}^{l \times d'}$, respectively. Finally, the commonsense knowledge is infused in the language representations using OKTE, generating the final vector $Z_g \in \mathbb{R}^{l \times d'}$. 

\paragraph{Information Fusion Module:}
Each of the modules mentioned above integrates salient information in the language representations. The information fusion module aggregates the knowledge obtained from all other modules using attention-based mutual interaction \cite{NIPS2017_3f5ee243}. Concretely, we concatenate the vectors of $Z$, $Z_m$, and $Z_g$ and pass them through a round of dimensionality reduction and non-linear transformations. Then, we use a self-attention mechanism so that the information obtained from each component interacts with one another. The final generated vector $Z'_c \in \mathbb{R}^{l \times d}$ is passed to a classifier with a softmax activation to predict the final labels.

\begin{table*}[t!]
\centering
\resizebox{0.9\textwidth}{!}{%
\begin{tabular}{cccccccccccccccccc}
\toprule
\multicolumn{2}{c}{\multirow{2}{*}{\textbf{Model Details}}} & \multicolumn{3}{c}{\textbf{Hero}}                                                           & \multicolumn{3}{c}{\textbf{Villain}}                                                        & \multicolumn{3}{c}{\textbf{Victim}}                                                         & \multicolumn{3}{c}{\textbf{Other}}                                                          & \multicolumn{3}{c}{\textbf{Macro-F1}}                                                       & \multirow{2}{*}{\textbf{Acc.}} \\ 
                                & & \multicolumn{1}{c}{\textbf{Prec}} & \multicolumn{1}{c}{\textbf{Rec}} & \textbf{F1} & \multicolumn{1}{c}{\textbf{Prec.}} & \multicolumn{1}{c}{\textbf{Rec.}} & \textbf{F1} & \multicolumn{1}{c}{\textbf{Prec.}} & \multicolumn{1}{c}{\textbf{Rec.}} & \textbf{F1} & \multicolumn{1}{c}{\textbf{Prec.}} & \multicolumn{1}{c}{\textbf{Rec.}} & \textbf{F1} & \multicolumn{1}{c}{\textbf{Prec.}} & \multicolumn{1}{c}{\textbf{Rec.}} & \textbf{F1} &                                    \\ \cmidrule(lr){1-2}\cmidrule(lr){3-5}\cmidrule(lr){6-8}\cmidrule(lr){9-11}\cmidrule(lr){12-14}\cmidrule(lr){15-17}\cmidrule(lr){18-18} 
\multirow{6}{*}{\rotatebox{90}{\textbf{Text-only}}} & Dis. BERT                  & \multicolumn{1}{c}{.154}              & \multicolumn{1}{c}{.115}           & .132       & \multicolumn{1}{c}{.431}              & \multicolumn{1}{c}{.400}           & .415       & \multicolumn{1}{c}{.292}              & \multicolumn{1}{c}{.246}           & .267       & \multicolumn{1}{c}{.856}              & \multicolumn{1}{c}{.881}           & .868       & \multicolumn{1}{c}{.433}              & \multicolumn{1}{c}{.411}           & .420       & .766                              \\ 
& BERT                        & \multicolumn{1}{c}{.250}              & \multicolumn{1}{c}{.115}           & .158       & \multicolumn{1}{c}{.484}              & \multicolumn{1}{c}{.446}           & .464       & \multicolumn{1}{c}{.358}              & \multicolumn{1}{c}{.254}           & .297       & \multicolumn{1}{c}{.864}              & \multicolumn{1}{c}{.905}           & .884       & \multicolumn{1}{c}{.489}              & \multicolumn{1}{c}{.430}           & .451       & .791                              \\ 
& RoBERTa                     & \multicolumn{1}{c}{.243}              & \multicolumn{1}{c}{.173}           & .202       & \multicolumn{1}{c}{.473}              & \multicolumn{1}{c}{.400}           & .433       & \multicolumn{1}{c}{.346}              & \multicolumn{1}{c}{.386}           & .365       & \multicolumn{1}{c}{.866}              & \multicolumn{1}{c}{.891}           & .879       & \multicolumn{1}{c}{.482}              & \multicolumn{1}{c}{.463}           & .470       & .782                              \\ 
& XLNet                       & \multicolumn{1}{c}{.172}              & \multicolumn{1}{c}{.096}           & .123       & \multicolumn{1}{c}{.491}              & \multicolumn{1}{c}{.386}           & .432       & \multicolumn{1}{c}{.383}              & \multicolumn{1}{c}{.272}           & .318       & \multicolumn{1}{c}{.852}              & \multicolumn{1}{c}{.910}           & .880       & \multicolumn{1}{c}{.475}              & \multicolumn{1}{c}{.416}           & .438       & .788                              \\ 
& DeBERTa                     & \multicolumn{1}{c}{.250}              & \multicolumn{1}{c}{.250}           & .250       & \multicolumn{1}{c}{.469}              & \multicolumn{1}{c}{\textbf{.591}}           & .523       & \multicolumn{1}{c}{.395}              & \multicolumn{1}{c}{.395}           & .395       & \multicolumn{1}{c}{\textbf{.889}}              & \multicolumn{1}{c}{.847}           & .868       & \multicolumn{1}{c}{.501}              & \multicolumn{1}{c}{\textbf{.521}}           & .509       & .776                              \\ 
& DeBERTa(l)                & \multicolumn{1}{c}{.268} & \multicolumn{1}{c}{.212} & \multicolumn{1}{c}{.237} & \multicolumn{1}{c}{\textbf{.604}} & \multicolumn{1}{c}{.440} & \multicolumn{1}{c}{.509} & \multicolumn{1}{c}{.543} & \multicolumn{1}{c}{\textbf{.500}} & \multicolumn{1}{c}{.521} & \multicolumn{1}{c}{.870} & \multicolumn{1}{c}{.922} & \multicolumn{1}{c}{\textbf{.895}} & \multicolumn{1}{c}{.518} & \multicolumn{1}{c}{.571} & \multicolumn{1}{c}{.540} & \multicolumn{1}{c}{\textbf{.818}}                                    \\ \hline

\multirow{5}{*}{\rotatebox{90}{\textbf{Vision-only}}} & ResNet                     & \multicolumn{1}{c}{0}                  & \multicolumn{1}{c}{0}               & 0           & \multicolumn{1}{c}{.467}              & \multicolumn{1}{c}{.303}           & .367       & \multicolumn{1}{c}{.667}              & \multicolumn{1}{c}{.053}           & .097       & \multicolumn{1}{c}{.827}              & \multicolumn{1}{c}{\textbf{.948}}           & .883       & \multicolumn{1}{c}{.490}              & \multicolumn{1}{c}{.326}           & .337       & .793                              \\ 
& ConvNeXT                 & \multicolumn{1}{c}{0}                  & \multicolumn{1}{c}{0}               & 0           & \multicolumn{1}{c}{.398}              & \multicolumn{1}{c}{.329}           & .360       & \multicolumn{1}{c}{.333}              & \multicolumn{1}{c}{.009}           & .017       & \multicolumn{1}{c}{.828}              & \multicolumn{1}{c}{.924}           & .873       & \multicolumn{1}{c}{.390}              & \multicolumn{1}{c}{.315}           & .313       & .776                              \\ 
& ViT                        & \multicolumn{1}{c}{0}                  & \multicolumn{1}{c}{0}               & 0           & \multicolumn{1}{c}{.436}              & \multicolumn{1}{c}{.380}           & .406       & \multicolumn{1}{c}{.250}              & \multicolumn{1}{c}{.009}           & .016       & \multicolumn{1}{c}{.836}              & \multicolumn{1}{c}{.926}           & .878       & \multicolumn{1}{c}{.380}              & \multicolumn{1}{c}{.329}           & .325       & .785                              \\ 
& BEiT                       & \multicolumn{1}{c}{0}                  & \multicolumn{1}{c}{0}               & 0           & \multicolumn{1}{c}{.413}              & \multicolumn{1}{c}{.420}           & .416       & \multicolumn{1}{c}{.500}              & \multicolumn{1}{c}{.009}           & .017       & \multicolumn{1}{c}{.838}              & \multicolumn{1}{c}{.907}           & .871       & \multicolumn{1}{c}{.438}              & \multicolumn{1}{c}{.334}           & .326       & .775                              \\ 
& SWIN                       & \multicolumn{1}{c}{0}                  & \multicolumn{1}{c}{0}               & 0           & \multicolumn{1}{c}{.393}              & \multicolumn{1}{c}{.326}           & .356       & \multicolumn{1}{c}{.231}              & \multicolumn{1}{c}{.026}           & .047       & \multicolumn{1}{c}{.828}              & \multicolumn{1}{c}{.919}           & .871       & \multicolumn{1}{c}{.363}              & \multicolumn{1}{c}{.318}           & .319       & .772                              \\ \hline
\multirow{7}{*}{\rotatebox{90}{\textbf{Multimodal}}} & CLIP                            & \multicolumn{1}{c}{0}                   & \multicolumn{1}{c}{0}                &       0      & \multicolumn{1}{c}{.478}                   & \multicolumn{1}{c}{.311}                &      .378       & \multicolumn{1}{c}{\textbf{.691}}                   & \multicolumn{1}{c}{.071}                &      .104       & \multicolumn{1}{c}{.845}                   & \multicolumn{1}{c}{.951}                &       .886      & \multicolumn{1}{c}{.393}                   & \multicolumn{1}{c}{.335}                &      .342       &                   .786                 \\ 
& ViLBERT                         & \multicolumn{1}{c}{.078}                   & \multicolumn{1}{c}{.034}                &     .045        & \multicolumn{1}{c}{.409}                   & \multicolumn{1}{c}{.513}                &      .469       & \multicolumn{1}{c}{.398}                   & \multicolumn{1}{c}{.276}                &      .321       & \multicolumn{1}{c}{.856}                   & \multicolumn{1}{c}{.843}                &       .849      & \multicolumn{1}{c}{.428}                   & \multicolumn{1}{c}{.408}                &      .421       &          .761                          \\ 
& V-BERT                      & \multicolumn{1}{c}{.143}                   & \multicolumn{1}{c}{.077}                &       .100      & \multicolumn{1}{c}{.466}                   & \multicolumn{1}{c}{.560}                &        .508     & \multicolumn{1}{c}{.452}                   & \multicolumn{1}{c}{.333}                &       .384      & \multicolumn{1}{c}{.886}                   & \multicolumn{1}{c}{.879}                &       .882      & \multicolumn{1}{c}{.487}                   & \multicolumn{1}{c}{.462}                &       .469      &         .790                           \\ 
& MMTrans.                   & \multicolumn{1}{c}{0}                   & \multicolumn{1}{c}{0}                &      0       & \multicolumn{1}{c}{.516}                   & \multicolumn{1}{c}{.477}                &      .496       & \multicolumn{1}{c}{.447}                   & \multicolumn{1}{c}{.303}                &      .361       & \multicolumn{1}{c}{.814}                   & \multicolumn{1}{c}{.878}                &        .845     & \multicolumn{1}{c}{.437}                   & \multicolumn{1}{c}{.392}                &       .405      &                .747                    \\ 
& MMBT                            & \multicolumn{1}{c}{.103}                   & \multicolumn{1}{c}{.058}                &      .074       & \multicolumn{1}{c}{.446}                   & \multicolumn{1}{c}{.537}                &       .487      & \multicolumn{1}{c}{.414}                   & \multicolumn{1}{c}{.298}                &      .347       & \multicolumn{1}{c}{.881}                   & \multicolumn{1}{c}{.872}                &      .877       & \multicolumn{1}{c}{.458}              & \multicolumn{1}{c}{.438}           & .447       &         .780 \\ \cline{2-18}
& \model & \multicolumn{1}{c}{\textbf{.444}} & \multicolumn{1}{c}{\textbf{.385}} & \multicolumn{1}{c}{\textbf{.412}} & \multicolumn{1}{c}{.553} & \multicolumn{1}{c}{.534} & \multicolumn{1}{c}{\textbf{.544}} & \multicolumn{1}{c}{.505} & \multicolumn{1}{c}{.456} & \multicolumn{1}{c}{\textbf{.479}} & \multicolumn{1}{c}{.883} & \multicolumn{1}{c}{.897} & \multicolumn{1}{c}{.890} & \multicolumn{1}{c}{\textbf{.568}} & \multicolumn{1}{c}{\textbf{.596}} & \multicolumn{1}{c}{\textbf{.581}} & \multicolumn{1}{c}{.813} \\ \cline{2-18}
& $\Delta_{(\model - \text{V-BERT})}$ & \multicolumn{1}{c}{\textcolor{blue}{.301$\uparrow$}} & \multicolumn{1}{c}{\textcolor{blue}{.308$\uparrow$}} & \multicolumn{1}{c}{\textcolor{blue}{.312$\uparrow$}} & \multicolumn{1}{c}{\textcolor{blue}{.087$\uparrow$}} & \multicolumn{1}{c}{\textcolor{red}{.026$\downarrow$}} & \multicolumn{1}{c}{\textcolor{blue}{.036$\uparrow$}} & \multicolumn{1}{c}{\textcolor{blue}{.053$\uparrow$}} & \multicolumn{1}{c}{\textcolor{blue}{.123$\uparrow$}} & \multicolumn{1}{c}{\textcolor{blue}{.095$\uparrow$}} & \multicolumn{1}{c}{\textcolor{red}{.003$\downarrow$}} & \multicolumn{1}{c}{\textcolor{blue}{.018$\uparrow$}} & \multicolumn{1}{c}{\textcolor{blue}{.008$\uparrow$}} & \multicolumn{1}{c}{\textcolor{blue}{.081$\uparrow$}} & \multicolumn{1}{c}{\textcolor{blue}{.134$\uparrow$}} & \multicolumn{1}{c}{\textcolor{blue}{.112$\uparrow$}} & \multicolumn{1}{c}{\textcolor{blue}{.023$\uparrow$}} \\ \hline
\end{tabular}%
}
\caption{Benchmarking results ($0$'s indicate no correct prediction). $\Delta_{(\texttt{X} - \texttt{Y})}$: Performance difference between models \texttt{X} and \texttt{Y}, \textcolor{blue}{$\#\uparrow$} and \textcolor{red}{$\#\downarrow$}: Absolute increment and decrement respectively.}
\label{tab:baseline}
\end{table*}

\section{Baselines}


    \paragraph{\textit{Unimodal Systems:}} We use a variety of text-based and image-based models as our baseline systems. Starting with the text baselines, we use {\bf BERT} \cite{devlin-etal-2019-bert}, and variants thereof such as {\bf DistilBERT} \cite{sanh2019distilbert}, {\bf RoBERTa} \cite{liu2019roberta}, {\bf XLNet} \cite{yang2019xlnet},  and {\bf DeBERTa} \cite{he2020deberta}. For the image-based baselines, we start with a representation based on {\bf ResNet-50} \cite{He_2016_CVPR}, followed by {\bf Vision Transformer (ViT)} \cite{dosovitskiy2020image} and {\bf SWIN} \cite{Liu_2021_ICCV}, which is hierarchical version of ViT that uses shifted windows. We also include the recently proposed {\bf ConvNeXT} \cite{Liu_2022_CVPR} and {\bf BEiT} \cite{bao2021beit}, which uses self-supervised pre-training of Vision Transformers.
    
    \paragraph{\textit{Multi-modal Systems:}} 
    We use various variants of multi-modal pre-trained systems from the MMF Framework.
    \textbf{MMF Transformer} is a library Transformer model that uses visual and language tokens with self-attention. \textbf{MMBT:} multi-modal Bi-transformer \cite{kiela2020supervised} captures the two modalities' intra-modal and inter-modal dynamics. \textbf{ViLBERT} is a vision and language BERT \cite{lu2019vilbert}, a strong model with the task-agnostic joint representation of images and text. \textbf{ViLBERT CC} is pre-trained on conceptual captions \cite{sharma2018conceptual} based pretext task. \textbf{Visual BERT}  \cite{li2019visualbert}, also pre-trained using MS COCO \cite{lin2014microsoft}, implicitly aligns the input text and regions in the input image using self-attention.
    CLIP \cite{radford2021learning} leverages image--text contrastive pretraining.

\section{Experiments}

\paragraph{\bf Experimental Details}

We present bench-marking results in Table~\ref{tab:baseline}, comparison with the shared-task submissions (c.f. Table \ref{tab:hvvcomp})
averaged over five independent runs, while for the ablation study in Table \ref{tab:ablation}, we compare the \textit{best} check-points for different \model\ component-wise evaluations to assess the performance bounds, on the \textit{diversified} test set.
We use 
precision, recall, and F1 for individual classes, and macro-averaged for the overall assessment. 
\footnote{See Appendix \ref{sec:hyperparameters} for additional experimental details.}


\paragraph{\bf Comparative Analysis.}

\noindent$\vartriangleright$ {\bf Unimodal Models:} Despite the evident glorification cues within the \textit{Hero} references in memes, large pre-trained models are observed to depict limitations. This can be observed from Table~\ref{tab:baseline}, wherein significantly low performance is observed for strong models like DistilBERT, BERT, and XLNet, with F1 scores of 0.132, 0.158, and 0.123, respectively. On the other hand, RoBERTa enhances the class-specific F1 score by about 0.05 absolute points, which suggests the efficacy the exhaustive hyperparameter tuning can induce. Finally, DeBERTa-based unimodal systems yield the highest optimal F1 scores among all unimodal models evaluated. The scores observed are 0.250 and 0.237 for the base and the large variants, respectively.  

For the \textit{Villain} category, BERT can be observed (c.f. Table~\ref{tab:baseline}) to yield relatively better performance as against Hero detection, with an F1 score of 0.464. This is in contrast to the sub-par performances by DistilBERT, RoBERTa, and XLNet, yielding 0.415, 0.433, and 0.432 F1 scores, respectively. As with the Hero detection, DeBERTa-base and large models outperform the other models for both text and vision modalities.

Interestingly, none of the text-only models, except for DeBERTa large, could beat the F1 score of 0.395 for DeBERTa-base. The lower performance is primarily due to inadequate categorical representation within the dataset. Moreover, the inherent complexity in distinguishing villains from victims confuses the model. The modelling efficacy solicited for such a scenario is suggested by over 0.10 absolute point enhancement by DeBERTa-large, which has 4X more backbone parameters than DeBERTa-base. 

\textit{Other} category, having the majority representation with over 6K unique references within memes, projects the highest F1 scores with an average of 0.879. Finally, the overall Macro-F1 scores reflect the category-wise trend observed, with DeBERTa large leading with an F1 score of 0.540. DeBERTa-base follows it with 0.509 and the rest with an average F1 score of 0.444.    

For intuitive reasons, the visual modality, not indicative of complex role semantics, yields poor performance compared to text-only models. As a result, none of the image-only models (ConvNext, ViTBEiT, and SWIN) are observed to make any headway in correctly detecting a Hero reference in memes (c.f. Table~\ref{tab:baseline}). At the same time, except for the Villain category, a simple ResNet-based model outperforms, albeit with fine margins, the rest of the models within the categories Victim and Other, with F1 scores of 0.097 and 0.883, respectively. This highlights the efficacy of global image representations against tokenized (or patched) ones for factoring visual features, especially where there is not much visual-linguistic grounding to be leveraged. On average, the image-only models can project a paltry F1 score of 0.324.

\begin{table}[t!]
\centering
\resizebox{\columnwidth}{!}{%
\begin{tabular}{@{}lcccccc@{}}
\toprule
\multicolumn{1}{c}{\textbf{Model Details}}           & \textbf{HER}  & \textbf{VIL}  & \textbf{VIC}  & \textbf{OTH}  & \textbf{F1}  & \textbf{Acc.}  \\ \cmidrule(lr){1-1}\cmidrule(lr){2-5}\cmidrule(lr){6-7}
Simple early-fusion (DeBERTa + ViT) & 0.31 & 0.55 & 0.50 & 0.88 & 0.56     & 0.79 \\
(a). $+$ Meme (image) caption              & 0.32 & 0.52 & 0.48 & 0.89 & 0.55     & 0.80 \\
(b). $+$ Face labels                       & 0.26 & 0.56 & 0.51 & 0.89 & 0.56     & 0.82 \\
(c). [(a) $+$ (b)] $+$ Commonsense KG       & 0.36 & 0.54 & 0.45 & 0.89 & 0.56     & 0.81 \\
(d). [(a) $+$ (b) $+$ (c)] + CAT & 0.28 & 0.53 & 0.49 & 0.89 & 0.55     & 0.81 \\
(e). [(a) $+$ (b) $+$ (c)] + OTK (\model)                                 & \textbf{0.38} & \textbf{0.57} & \textbf{0.53} & \textbf{0.90} & \textbf{0.60}     & \textbf{0.83} \\ \bottomrule
\end{tabular}%
}
\caption{Ablation study: Comparing \model\ and its sub-modules over \textit{best} model results. CAT: Fusion via concatenation, OTK: Fusion via Optimal Transport Kernel, KG: Knowledge Graph.} 

\label{tab:ablation}
\end{table}
$\vartriangleright$ {\bf Multi-modal Models.} Several state-of-the-art multimodal systems, on average, are observed to yield a Macro-F1 score of 0.416, which lies between the text-only (0.444) and the image-only (0.324) models. Besides the category-wise performance trend being similar to the one observed for unimodal models, multimodal systems like VisualBERT and MMBT yield the highest Macro-F1 scores of 0.468 and 0.447, respectively. This is likely due to the joint attentive modelling, multimodal pre-training using standard datasets like MS COCO, and fine-tuning adopted by these models. Other competitive models like CLIP, ViLBERT, and MM Transformer achieve Macro-F1 scores of up to 0.342, 0.421, and 0.405, respectively. This either suggests that the image component induces additional noise within the models or that the existing multimodal systems do not effectively capture the complex pragmatics that semantic role-labeling in memes solicits. The former is less likely, as intuitively, memetic visuals do provide minor yet decisive semantics toward holistic assimilation of the meme's message. 
Our proposed approach \model, is observed to address the required cross-modal association by achieving impressive F1 scores across different roles and a 0.581 Macro-F1 score, which induces an enhancement of almost 4\% and 12 \% over DeBERTa large (unimodal best) and VisualBERT (multimodal best), respectively.

\paragraph{Ablations Results.}
 Table \ref{tab:ablation} depicts ablation results, wherein the basic early-fusion setup involving DeBERTa and ViT performs well for \textit{villain} category with $0.55$ F1 score. Adding meme-image captions enhances the \textit{hero} predictions and overall accuracy marginally by $1\%$ each. Face labels, although, significantly enhance the prediction of \textit{villain} and \textit{victim}, suggesting lexical and semantic utility via \textit{face labels}. The overall performance remains unchanged due to compromised \textit{hero} predictions. Besides yielding balanced scores, adding a commonsense module elevates hero prediction distinctly to $0.36$, effectively indicating its utility, especially for the under-represented role category. Finally, \model\ with OTK-based embedding yields optimal cross-modal correspondence, as observed from the best performances across roles and metrics evaluated, showcasing the constituting feature's utility towards addressing the given task. Although \textit{not} averaged over multiple runs, the \textit{best} model check-point for \model\ is observed to yield an impressive overall macro-F1 score of 0.60, which is the best score observed across experiments, suggesting an upper-bound for \model's performance.

\begin{table}[t!]
\centering
\resizebox{\columnwidth}{!}{%
\begin{tabular}{ccccc}
\toprule
\textbf{Rank} & \textbf{System}       & \textbf{Prec.} & \textbf{Rec.} & \textbf{F1} \\ \midrule
-             & \model & 0.568              & 0.596           & 0.581       \\ 
1             & Logically \citep{kun-etal-2022-logically}            & 0.544              & 0.610           & 0.571       \\ 
2             & c1pher \citep{c1pher:2022}                & 0.527              & 0.581           & 0.547       \\ 
3             & smontariol \citep{smontariol:2022}           & 0.580              & 0.450           & 0.485       \\ 
4             & zhouziming \citep{DD-TIG:2022}            & 0.480              & 0.450           & 0.462       \\ 
5             & IIITDWD \citep{fharook:2022}              & 0.256              & 0.238           & 0.239       \\ 
6             & rabindra.nath \citep{rabindranath:2022}        & 0.253              & 0.253           & 0.237       \\ \bottomrule
\end{tabular}%
}
\caption{Comparison to the results on the CONSTRAINT@ACL-2022 shared-task on \dataset.}
\label{tab:hvvcomp}
\end{table}
\paragraph{Comparison to Previous Work.}
\label{sec:comparison}

Table~\ref{tab:hvvcomp} showcases the best-performing systems from CONSTRAINT@ACL-2022's shared-task. Since all of the shared-task submissions used ensemble techniques of multiple models, we present their best individual model results for a fair comparison with our proposed model: \model. The macro-F1 score varies by 0.334 points across the participants, emphasizing model selection. We can draw parallels between Table~\ref{tab:baseline} and Table~\ref{tab:hvvcomp} with Logically, c1pher, smontariol, and zhouziming using DeBERTa, RoBERTa, VisualBERT, and VisualBERT, respectively. They exhibit marginal improvements using loss-weighting techniques and additional classification layers. Our model \model\ performs more consistently, especially with the class-wise scores of \emph{Hero} (c.f. Table~\ref{tab:baseline}), establishing its efficacy across the category types and limited categorical data representation. Besides class-wise consistency, \model\ outperforms other shared-task submissions as a stand-alone system. Also, despite being relatively complex, \model\ produces consistent class-wise and better overall performance, suggesting marginal yet robust modelling capacity facilitated by \model's vision + common-sense module's interaction with the textual signals. 

\begin{figure}[t!]
    \centering
    \includegraphics[width=\columnwidth]{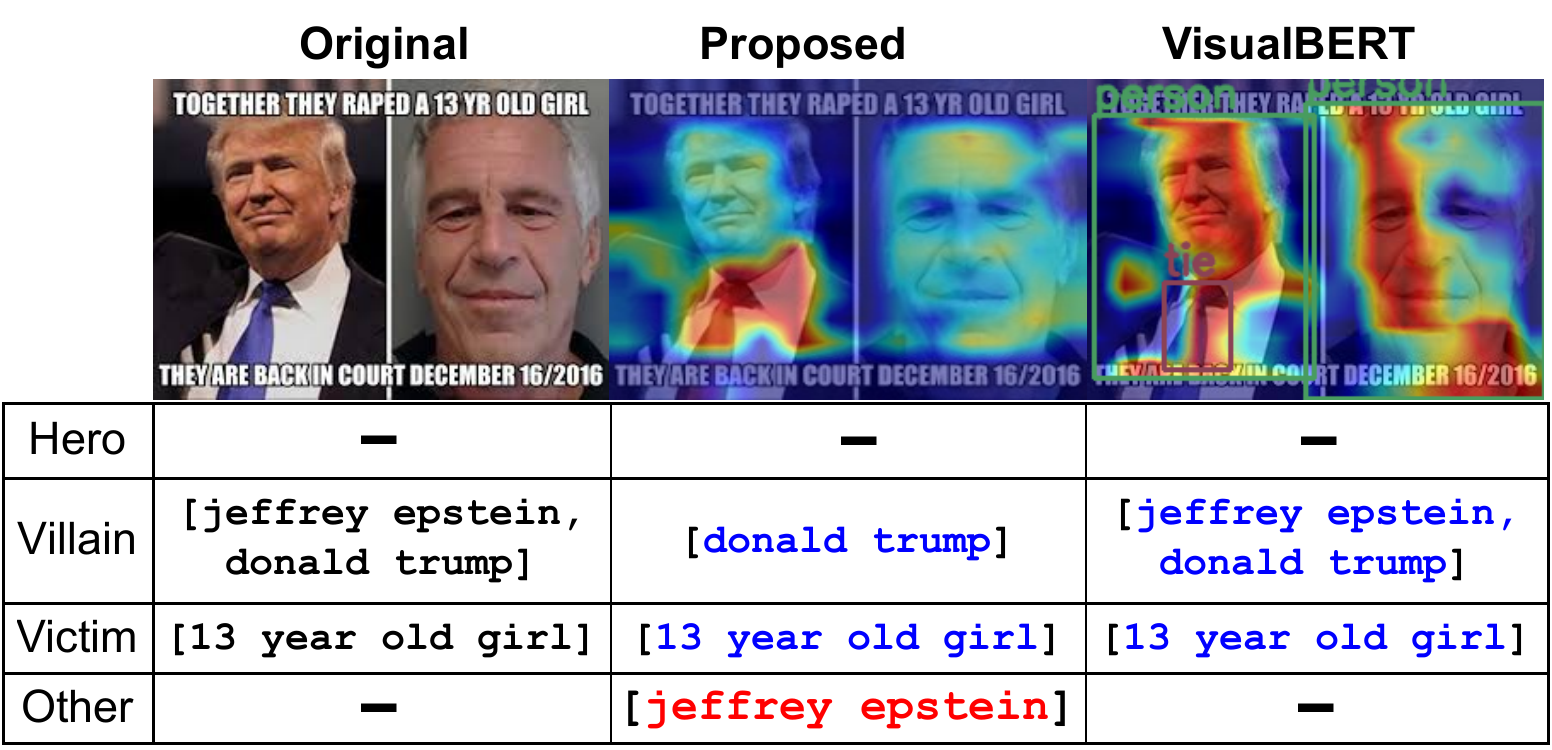}
    \caption{Error analysis for \model\ and VisualBERT.}
    \label{fig:err_analysis}
\end{figure}

\paragraph{Statistical Significance.}
We performed a bootstrapping significance test w.r.t the proposed model (\model), and the previous best solution by team Logically \cite{kun-etal-2022-logically} via random sampling with replacement strategy, with N=1000 over 1000 simulations, and observed a `p-value' of $0.0410$. This indicates a subtle yet encouraging confidence margin in the model predictions. This could be likely due to better predictions across the four categories, including `hero', which most of the other models compared are empirically observed to struggle at. These aspects corroborate the semantic-role label-agnostic characteristics of the proposed model.

\paragraph{Error Analysis.}
The example depicted in Fig. \ref{fig:err_analysis} insinuates `donald trump' and `jeffrey Epstein's as \textit{villains} while \textit{victimizes} a `13 year old girl'. Visually, there isn't much to consider towards adjudicating the former two entities as villains, except the expressions of `jeffrey Epstein's exuding somewhat \textit{sinister} looks. Whereas \textit{vilifying} connotation is implied primarily by the embedded text. Now, although \model\ predicts the roles of `donald trump'  and  `13 year old girl' correctly as \textit{villain} and \textit{victim}, respectively, it fails to detect `jeffrey epstein as a \textit{villain} and categorizes it as an \textit{other}. This example highlights the limitations of \model\ in terms of its inherent modality-specific biases. A ViT-based image encoder, due to its disparate patch-wise processing, and self-attention across the input patches, leads to noisy visual attention. On the other hand, VisualBert-based predictions replete with pre-trained common-sense semantics are better positioned for this case to capture the required semantic indicators, a portion of \textit{epstein's} facial expression in this case. Also, as the dataset is well-stocked with examples where `Donald trump' is \textit{vilified},
both models being compared assign the role of \textit{villain} to `donald trump'. 

\begin{figure}[t!]
    \centering
    \includegraphics[width=0.8\columnwidth]{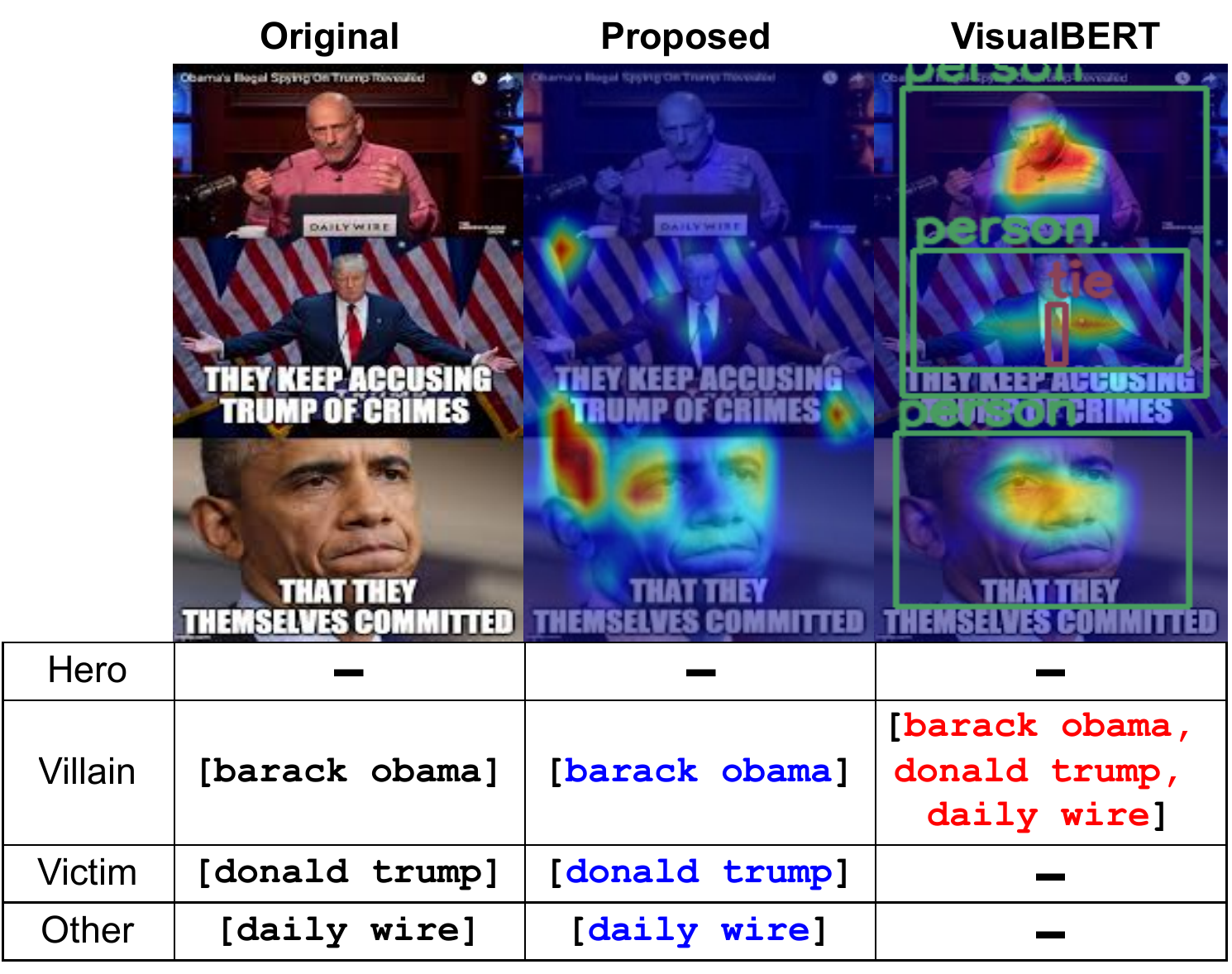}
    \caption{Interpretability analysis for \model; proposed model and VisualBERT; best multimodal baseline.}
    \label{fig:int_analysis}
\end{figure}

\section{\bf Interpretability}
\label{sec:app:int}
The predictive capacity of \model\ can be assessed by comparing its interpretability with that of VisualBERT, the best-performing baseline. We investigate a reasonably complex example depicted in Fig \ref{fig:int_analysis}, for which \model\ accurately predicts all the entity roles, but VisualBERT wrongly predicts all of them, as \textit{villains}. Attention map for \model\ reveals that it primarily attends to the face of `barack obama', while leveraging other contextual cues, likely via the common-sense and visual description-based signals, leading to a correct assignment of roles. On the other hand, VisualBERT, having a Faster R-CNN-based image encoder, primarily pre-trained for the common object-detection task, fails to uniquely attend to different semantically referred entities within the meme image and additionally adds noisy features related to objects like a \textit{tie}, without any additional contextual knowledge. This validates \model's superior discriminatory capacity and interpretability over VisualBERT.

\section{\bf Generalizability}
\label{sec:app:gen}
Since, the test set 
consists of all unique memes, there is a higher chance that there are many entities as part of it that are not seen during the training stage. Table \ref{tab:baseline} attests to the \model's potential to succeed across different roles. This suggests that \model\ can adapt to the domain-specific nuances offered by the variety and complexity associated with visual-semantic roles within memes. Through the results observed in this work, we attempt to highlight that existing \textit{standard} approaches either have inconsistent performances across the roles: hero, villain, victim, and other, or yield low scores for particular role categories (c.f. Table \ref{tab:baseline}), which corroborates their limitations. On the other hand, \model\ yields a balanced performance and generalizes reasonably well for the least represented class in the dataset (\textit{hero}).

\section{Conclusion}
This paper addresses a recently proposed task of identifying the roles of entities in harmful memes and discusses its challenges. 
We further presented numerous unimodal and multi-modal baselines to benchmark \dataset. 
Moreover, we proffer \model, a contextual knowledge-enriched multi-modal framework that bolsters the multi-modal representations with entity-based external knowledge using a cross-modal attention scheme. \model\ shows noteworthy improvements over the baselines, thus justifying contextual knowledge inclusion. As for future investigations, we plan to conceive a more symbolic system with graph-based entity linking, commonsense knowledge, and visual concepts.


\section{Limitations}
\label{sec:app:limit}
As noted in the discussion dedicated to \textit{Error Analysis}, several entities tend to dominate specific roles within the dataset due to the realistic representation of the harmful referencing in memes. This not only biases the model against their generalizability but also poses challenges towards modelling entity-independent role detection hypotheses for a diverse set of entities. This especially calls for building models regularized to address such biases and more participatory initiatives toward curating better and more large-scale datasets.

\section{Ethics and Broader Impact}
\label{sec:app:ethics}
\paragraph{Reproducibility.}  We present detailed hyper-parameter configurations in Appendix \ref{sec:hyperparameters}. 

\paragraph{User Privacy.} 
The information depicted/used does not include any personal information. Copyright aspects are attributed to the dataset source.  




\paragraph{Biases.}
As per the authors, any biases found in the dataset are unintentional \cite{sharma2022findings}, and by conducting the study on this dataset we do not intend to cause harm to any group or individual. We acknowledge that detecting harmfulness can be subjective, and thus it is inevitable that there would be biases in gold-labelled data or in the label distribution. This is addressed by the dataset curators by using general keywords about US Politics, and also by following a well-defined schema, which sets explicit definitions for annotation. 

\paragraph{Misuse Potential.}
Our approach can be potentially used for ill-intended purposes, such as biased targeting of individuals/communities/organizations, etc. that may or may not be related to demographics and other information within the text. Intervention with human moderation would be required to ensure that this does not occur.


\paragraph{Intended Use.}

We make use of the existing dataset in our work in line with the intended usage prescribed by its creators and solely for research purposes. This applies in its entirety to its further usage as well. We do not claim any rights to the dataset used or any part thereof. We believe that it represents a useful resource when used appropriately.

\paragraph{Environmental Impact.}

Finally, large-scale models require a lot of computations, which contribute to global warming \cite{strubell2019energy}. However, in our case, we do not train such models from scratch; rather, we fine-tune them on a relatively small dataset.

\section*{Acknowledgements}
The work was supported by Wipro research grant.

\bibliography{eacl2023}
\bibliographystyle{acl_natbib}
\clearpage
\appendix

\section{Implementation Details and Hyperparameter Values} 
\label{sec:hyperparameters}

\begin{table}[t!]
\centering
\resizebox{1\columnwidth}{!}
{
\begin{tabular}{c  l  c  c  c  c  c  c }
\hline
\textbf{Modality} & \textbf{Model} &  \bf BS & \bf \#Epochs & \bf LR & \bf V-Enc & \bf T-Enc & \bf \#Param \\
\hline

\multirow{11}{*}{\centering \bf UM} & BERT & 8 & 15 & 1e-5 & - & \texttt{bert} & 25M\\
& DistilBERT & 8 & 15 & 1e-5 & - & \texttt{distilbert-base} & 66M\\
& XLNet & 8 & 15 & 1e-5 & - & \texttt{xlnet} & 116M\\
& RoBERTa & 8 & 15 & 1e-5 & - & \texttt{roberta-base} & 123M\\
& DeBERTa & 8 & 15 & 1e-5 & - & \texttt{deberta-base} & 86M\\
& DeBERTa-Large & 8 & 15 & 1e-5 & - & \texttt{deberta-large} & 304M\\
\cdashline{2-8}
& ResNet & 8 & 15 & 1e-5 & \texttt{resnet} & - & 25M\\
& ConvNeXT & 8 & 15 & 1e-5 & \texttt{convnet} & - & 50M\\
& ViT & 8 & 15 & 1e-5 & \texttt{vit} & - & 86M\\
& SWIN & 8 & 15 & 1e-5 & \texttt{swin} & - & 88M\\
& BEiT & 8 & 15 & 1e-51 & \texttt{beit} & - & 71M\\
\hline
\multirow{5}{*}{ \centering \bf MM} & MMFT & 16 & 20 & 0.001 & ResNet-152 & \texttt{bert} & 170M\\
& CLIP & 16 & 20 & 0.0001 & ViT & \texttt{clip} & 149M\\
& MMBT & 16 & 20 & 0.0001 & ResNet-152 & \texttt{bert} & 169M\\
& ViLBERT* & 16 & 10 & 0.0001 & Faster RCNN & \texttt{bert} & 112M\\
& V-BERT* & 16 & 10 & 0.0001 & Faster RCNN & \texttt{bert} & 247M\\ \cline{2-8}
& \model & 8 & 15 & 1e-5 & \texttt{vit} & \texttt{deberta-large} & 123M \\
\hline
\end{tabular}}
\caption{Hyperparameters summary. [BS$\rightarrow$Batch Size; LR$\rightarrow$Learning Rate; V/T-Enc$\rightarrow$Vision/Text-Encoder; \texttt{vit}$\rightarrow$\texttt{vit-base-patch16-224-in21k}; \texttt{bert}:$\rightarrow$\texttt{bert-base-uncased}; \texttt{xlnet}$\rightarrow$\texttt{xlnet-base-uncased};
\texttt{resnet}$\rightarrow$\texttt{resnet50}].}
\label{tab:hyperparameters}
\end{table}
We train all the models using PyTorch on an actively dedicated NVIDIA Tesla V100 GPU, with 32 GB dedicated memory, CUDA-11.2, and cuDNN-8.1.1 installed. For the unimodal models, we import all the pre-trained weights from the \texttt{TORCHVISION.MODELS}\footnote{\url{http://pytorch.org/docs/stable/torchvision/models.html}}, a sub-package of the PyTorch framework. We randomly initialise the remaining weights. 
\citet{sharma2022findings} re-annotate the HarMeme dataset \cite{pramanick-etal-2021-momenta-multimodal} by collecting annotator responses for different roles different entities within memes take. The re-annotated memes may or may not have harmful implications, contrary to the distinction modelled as part of original curation. However, they portray various entities within different contexts implying glorification, vilification, and victimisation.
\begin{wrapfigure}[8]{R}{0.20\textwidth}
\vspace{-20pt}
\centering
\includegraphics[width=\textwidth]{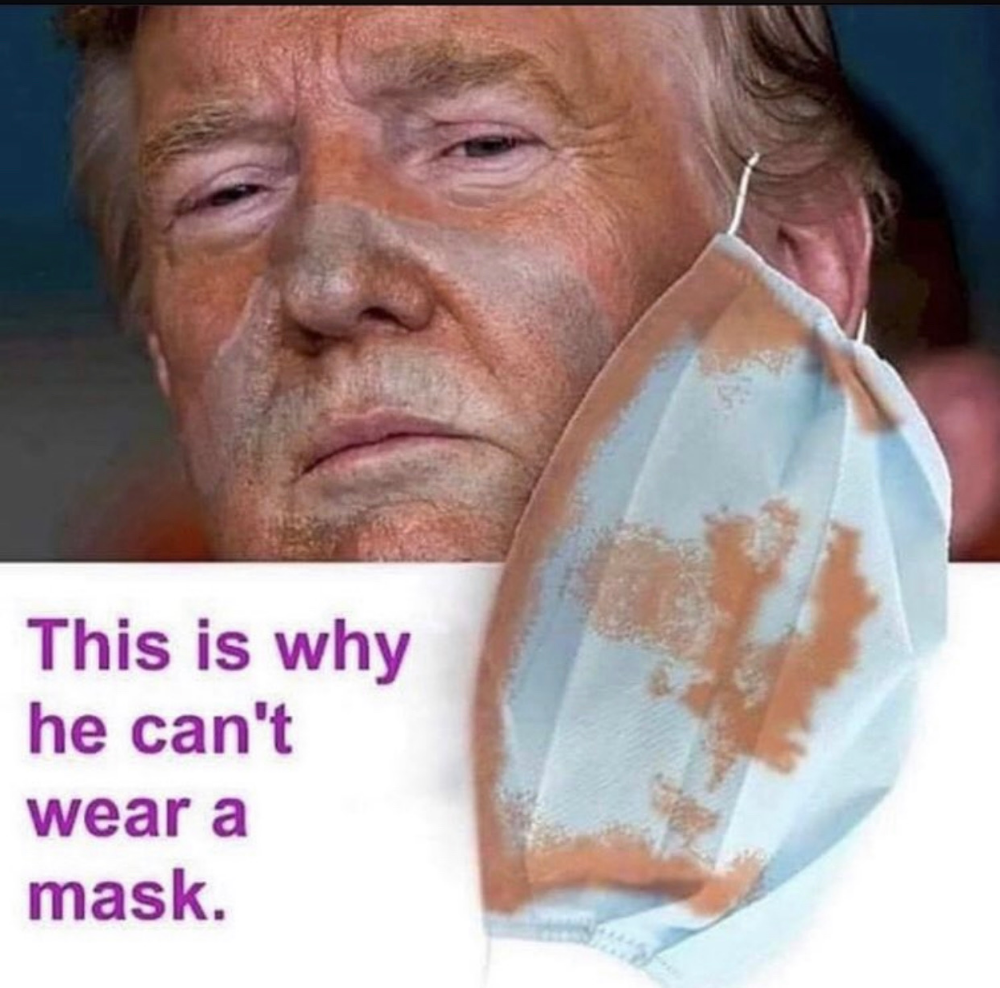}
\caption{\label{fig:donald_mask}Edge case}
\end{wrapfigure}
For most of our experiments, we use Adam optimiser \cite{kingma2014adam} with a learning rate of $1e^{-4}$ or $1e^{-5}$, a weight decay of $1e^{-5}$ and a Cross-Entropy (CE) loss as the objective function. We optimized our models to obtain hyper-parameter settings (c.f. Table \ref{tab:hyperparameters}) and early-stop to preserve our best state convergence. On average, it took approx. 2:30 hours to train a typical multi-modal neural model on a dedicated GPU system. 


\section{Additional details of \dataset}
\label{sec:app:datasetplus}


\subsection{\bf Edge Case}
Most memes are intended to project harmless mockery toward various sections of society. Such memes do not imply heroes, villains, or victims; instead, they disseminate harmless humour and trivial opinions. Therefore, \citet{sharma2022findings} do not presume any implications regarding these connotations and categorise them as `other', a fourth \textit{neutral} category, unless expressed otherwise in the meme.
A depiction of such a scenario can be observed in Fig. \ref{fig:donald_mask}. In this meme, it is unclear if Donald Trump is being vilified for being reluctant to use a mask or if the meme expresses a benign attempt at mocking his physical appearance with sarcasm. Additionally, since no background information would suffice to facilitate its complete assimilation, it is categorised as \textit{other}.

\begin{figure}[t!]
    \centering
    \includegraphics[width=\columnwidth]{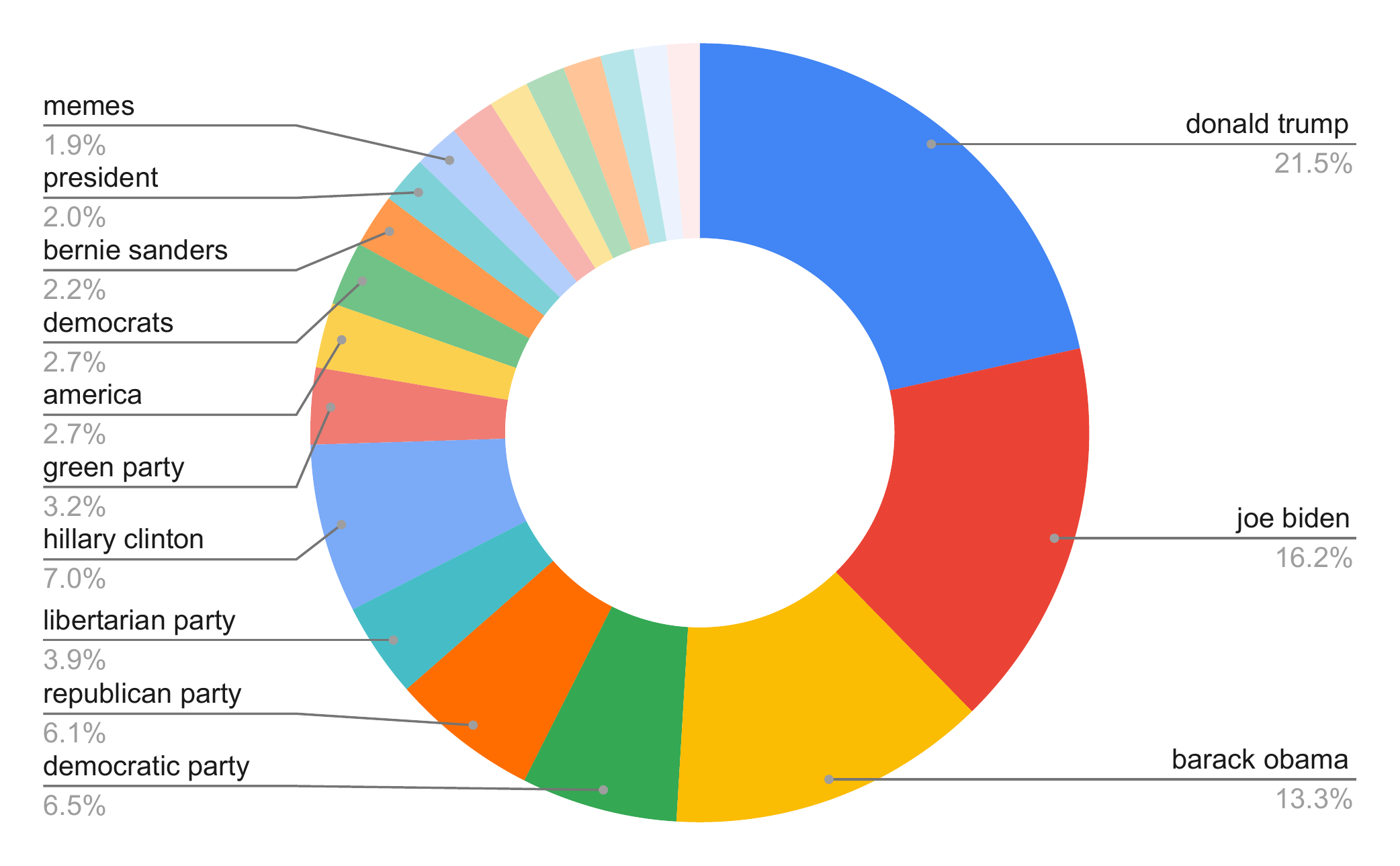}
    \caption{Entity distribution - US Politics.}
    \label{fig:count_pol}
\end{figure}
\begin{figure}[t!]
    \centering
    \includegraphics[width=\columnwidth]{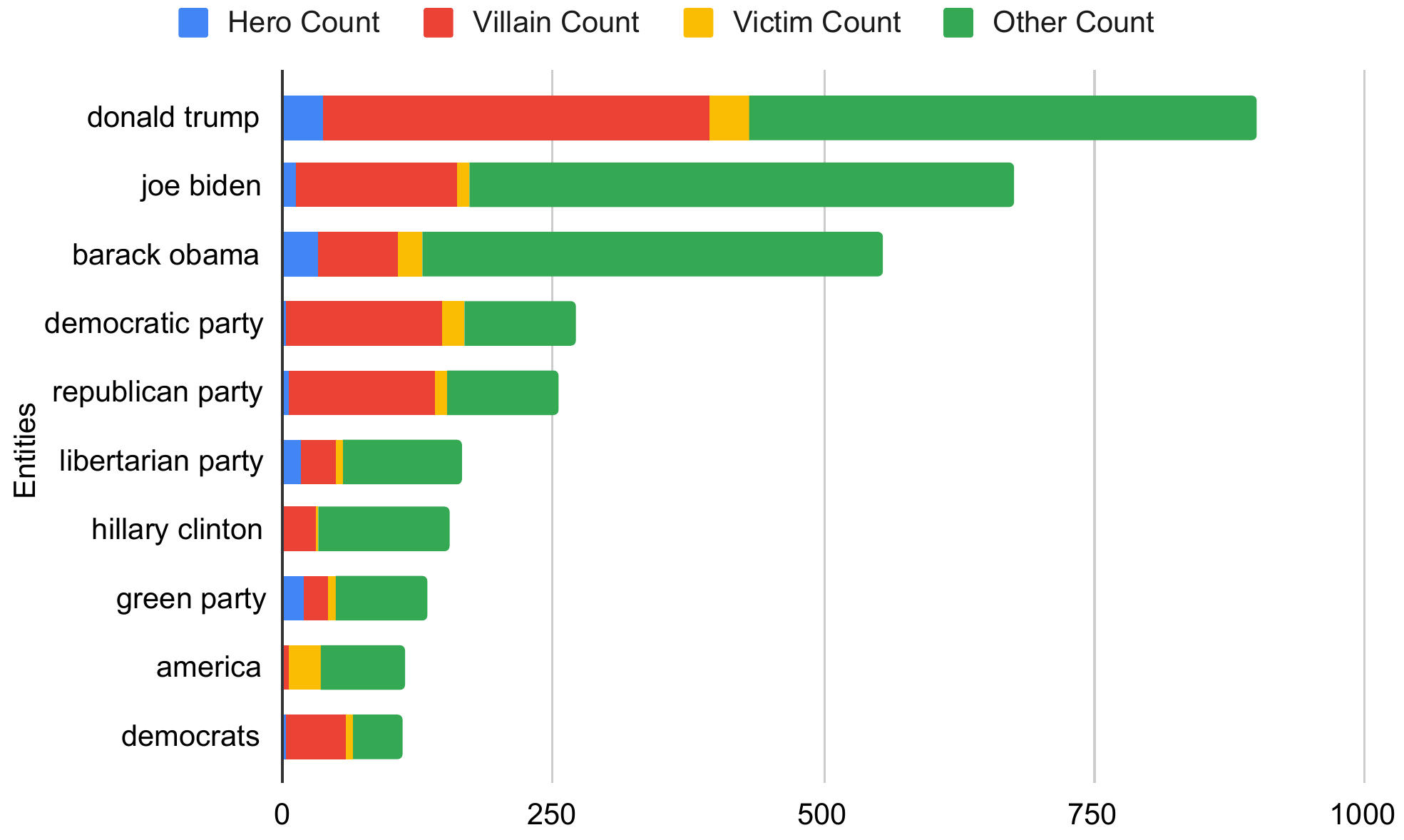}
    \caption{Entity-role distribution - US Politics.}
    \label{fig:dist_uspol}
\end{figure}

\subsection{\bf Analysing Different Entities and Roles}
Regardless of the connotations and domains in which various entities are referred, only a handful of entities/topics dominate pivotal referencing in memes. 
The entities in \textit{COVID-19} memes predominantly focus on \textit{Coronavirus}, {\em Donald Trump}, {\em mask}, {\em COVID-19}, and {\em work from home}. The entities of \textit{Donald Trump}, {\em Joe Biden}, {\em Barack Obama}, {\em Democratic Party}, and {\em Republican Party} crowd the US Politics memes, as shown in Fig. \ref{fig:count_pol}. Such trends highlight the key figures and real-world topics that dominated memetic communication on social media during the period this dataset was compiled, which coincided with the onset of the global COVID-19 pandemic and contemporary US Politics. 
In COVID-19 memes, we observe that entities like \textit{Donald Trump} and {\em China} are referenced almost equally within the vilifying and neutral contexts.
In contrast, entities like \textit{Corona beer}, {\em introverts}, and {\em Tom Hanks},  are invariably referenced in neutral contexts through irony, satire, or benign humour. Similar trends are observed in US Politics memes (Fig. \ref{fig:dist_uspol}), wherein entities like \textit{Donald Trump}, {\em Democratic Party}, {\em Republican Party}, and {\em Democrats} observe almost equivalent referencing as \textit{villain} and \textit{others}. Interestingly, as shown in Fig. \ref{fig:dist_uspol}, US Politics related memes depict a higher propensity towards vilification than that of COVID-19, as most of the prominent entities encountered are vilified at least once in the dataset.

\begin{figure}[t!]
    \centering
    \includegraphics[width=\columnwidth]{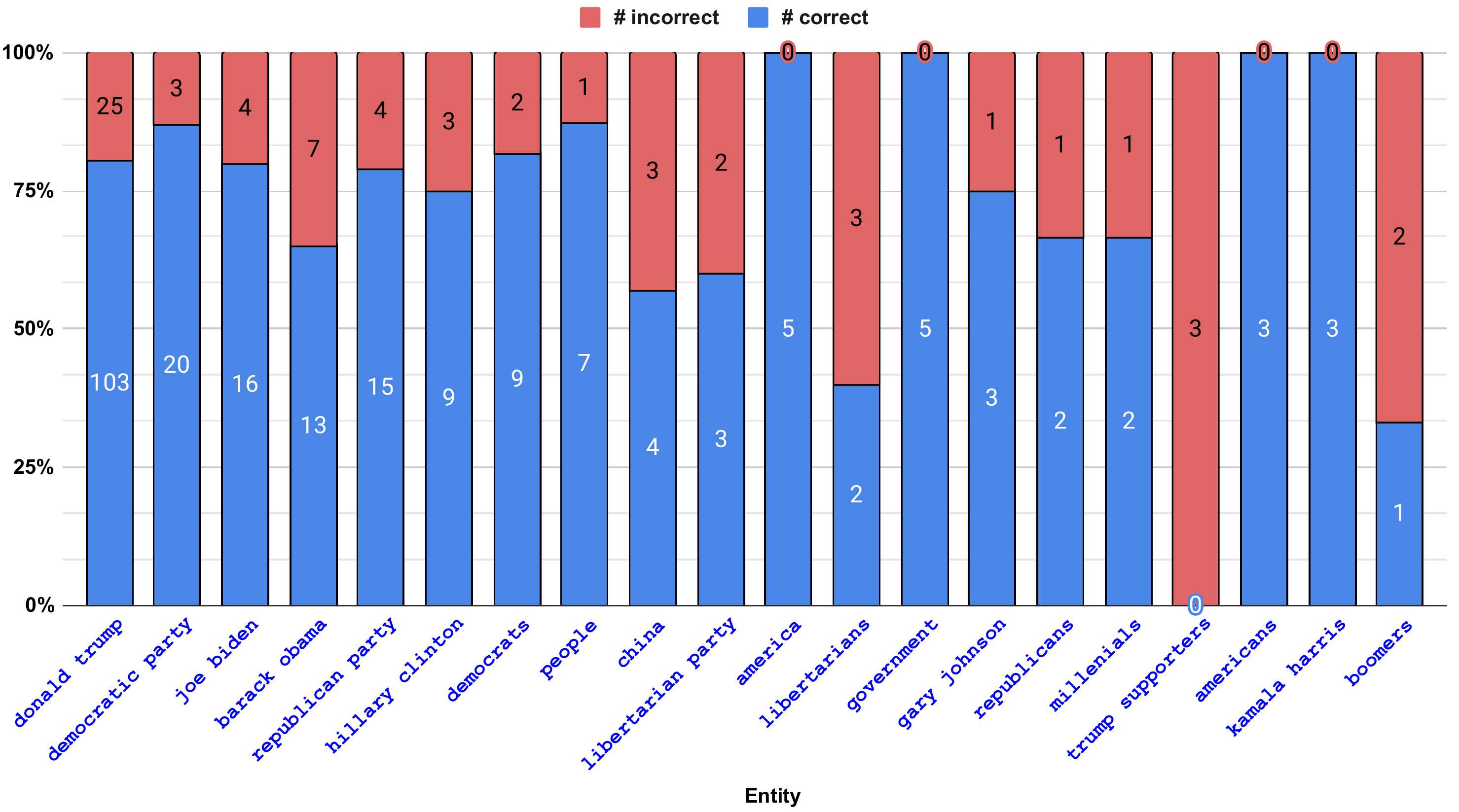}
    \caption{Entity role prediction summary for \dataset's (combined) test set. X-axis: Top 20 entities referenced; Y-axis: True positive rate (Recall). Values depicted at the center of each portion for each bar depicts corresponding total counts.}
    \label{fig:ent_analysis}
\end{figure}

\section{Entity Role Prediction Analysis.}
\label{sec:app:entpred}
Careful analysis of the role predictions for various entities elicits the correlation between the role-wise distribution and the test set predictions for different entities. This correlation can be observed from Fig. \ref{fig:ent_analysis}, wherein entities like \textit{Donald Trump, Democratic Party, Joe Biden, Republican Party, Democrats}, etc., which have a \textit{true positive rate} (recall) of atleast $75\%$, are specifically the ones that have relatively balanced role-wise distribution for the roles of \textit{villain} and \textit{other} within \dataset,\ (see Fig. 4 (main content). Interestingly, a lower but role-wise balanced representation within \dataset, does not appear to deter \model\ from yielding an impressive recall of $90\%$ for an entity such as \textit{Democrats}, which have a total of approximately $125$ samples in the training set. On the other hand, for entities like \textit{Barack Obama, China, Libertarian Party, Libertarian}, etc., the memetic portrayal is significantly skewed-in as an \textit{other}, suggesting distinct role-wise imbalance within memes that \model\ failed to accommodate, highlighting its limitations. Further, there are entities like \textit{America, Government, Americans, Kamala Harris}, etc., that register a $100\%$ recall. Role-wise distributions for such cases highlight a distinct majority of neutral connotations within both training and test splits via \textit{other} category, essentially suggesting possible biases within the role-prediction modelling setup. Entity \textit{Trump Supporters} is also observed to be present within the training set via another similar referencing \textit{Donald Trump Supporters}, which has a 1:5 ratio of villain/victim:other. This is complemented by the predominant yet balanced referencing of an independent entity \textit{Donald Trump}. This could lead to the model being confused for such entities as is depicted in Fig. \ref{fig:ent_analysis}.

\end{document}